\documentclass[journal]{IEEEtran}

\ifCLASSOPTIONcompsoc
  \usepackage[nocompress]{cite}
\else
  \usepackage{cite}
\fi

\ifCLASSINFOpdf

\else

\fi

\usepackage{footnote}
\usepackage{times}
\usepackage{epsfig}
\usepackage{graphicx}
\usepackage{amsmath}
\usepackage{amssymb}
\usepackage{algorithm}
\usepackage{algpseudocode}
\usepackage{booktabs}
\usepackage{multirow}
\usepackage{subfigure}

\graphicspath{{Figs/}}

\begin{document}

\title{Towards the Design of an End-to-End Automated System for Image and Video-based Recognition}



%

\author{Rama Chellappa$^1$, Jun-Cheng Chen$^3$, Rajeev Ranjan$^1$, Swami Sankaranarayanan$^1$, Amit Kumar$^1$,\\ Vishal M. Patel$^2$ and Carlos D. Castillo$^4$ 
\thanks{$^1$Department of ECE and the Center for Automation Research, UMIACS, University of Maryland, College Park, MD USA. email: \{rama,rranjan1,swamiviv,akumar14,carlos\}@umiacs.umd.edu}
\thanks{$^2$Department of Electrical and Computer Engineering at Rutgers University,
   Piscataway, NJ USA. email: vishal.m.patel@rutgers.edu.}    
\thanks{$^3$ Department of Computer Science and the Center for Automation Research, UMIACS, University of Maryland, College Park, MD,
USA. email: pullpull@cs.umd.edu.}    
\thanks{$^4$ Center for automation Research, UMIACS, University of Maryland, College Park, MD,
USA.} 

}


\maketitle

\begin{abstract}
Over many decades, researchers working in object recognition have
longed for an end-to-end automated system that will simply accept 2D
or 3D image or videos as inputs and output the labels of objects in
the input data. Computer vision methods that use representations
derived based on geometric, radiometric and neural considerations
and statistical and structural matchers and artificial neural
network-based methods where a multi-layer network learns the mapping
from inputs to class labels have provided competing approaches for
image recognition problems. Over the last four years, methods based
on Deep Convolutional Neural Networks (DCNNs) have shown impressive
performance improvements on object detection/recognition challenge
problems. This has been made possible due to the availability of
large annotated data, a better understanding of the non-linear
mapping between image and class labels as well as the affordability
of GPUs. In this paper, we present a brief history of developments
in computer vision and artificial neural networks over the last
forty years for the problem of image-based recognition. We then
present the design details of a deep learning system for end-to-end
unconstrained face verification/recognition. Some open issues
regarding DCNNs for object recognition problems are then discussed.
We caution the readers that the views expressed in this paper are
from the authors and authors only!
\end{abstract}

\IEEEpeerreviewmaketitle

\section{Introduction}
Over many decades, researchers working in object recognition have
longed for an end-to-end automated system that will simply accept 2D
or 3D image or videos as inputs and output the labels of objects in
the input data. In object recognition systems developed thus far,
representations such as templates, interest points, curves,
surfaces, appearance models, parts, histogram of gradients, scale
invariant feature transform, stochastic models, dynamic textures and
many others have been used. For recognition, statistical, syntactic
recognition methods, SVMs, graph matchers, interpretation trees, and
many others have been employed. Despite significant progress, the
performance of these systems has not been adequate for deployment.
In a parallel universe, systems based on artificial neural networks
have shown much promise since the mid-eighties. Given the
availability of millions of annotated data, GPUs and a better
understanding of the non-linearities, DCNNs are providing much
better performance on tasks such as object/face detection, object
recognition, face verification/recognition than traditional computer
vision methods.

In this paper, we briefly trace the history of development of object
recognition methods in images and videos using methods based on
traditional computer vision and artificial neural networks and then
present a case study in designing an end-to-end face
verification/recognition system using deep learning networks. While
the performance of methods based on DCNNs is impressive, we feel
that current day DCNNs are more like the Model-T cars. We conclude
the paper with a summary of remaining challenges to be addressed so
that deep learning networks can morph into high performance robust
recognition systems.

The organization of this paper is as follows: Section~\ref{sec:history} presents
a brief history of developments in computer vision and artificial neural
networks over the last forty years. Section~\ref{sec:janus} presents the design
details of a deep learning system for end-to-end unconstrained face
verification/recognition under the support of IARPA JANUS program. Some open
issues regarding DCNNs for object recognition problems are discussed in
Section~\ref{sec:con}.

\section{A Brief History of Developments in Image Recognition Using Computer
Vision and Neural Network-Based Methods}\label{sec:history}

Since the early sixties, when Robert's edge operator was introduced,
computer vision researchers have been working on designing various
object recognition systems. The goal has been to design an
end-to-end automated system that will simply accept 2D, 3D or video
inputs and spit out the class labels or identities of objects.
Beginning with template matching approaches in the sixties and
seventies, methods based on global and local shape descriptors were
developed. In the seventies, methods based on representations such
as Fourier descriptors, moments, Markov models, and statistical
pattern recognizers were developed. Even in the early years, the
need for making the global recognition approaches be invariant to
various transformations such as scale, rotation, etc. were
recognized and appropriate transformations were suggested in the
feature extraction stage; an alternative approach was to search over
these transformations during the classification step. Unlike these
global descriptors, local descriptors based on primitives such as
line segments, arcs etc. were used in either structural or syntactic
pattern recognition engines. For example, generative grammars of
various types were designed to parse the given object contour into
one of many classes. More information on these developments can be
found in \cite{fu_book, Fukunaga_book}.

In the eighties, statistical pattern recognition methods were seen
as not being able to handle occlusions or geometric representations.
Graph matching or relaxation approaches became popular for
addressing problems such as partial object matching. In the
mid-eighties, 3D range data of objects became available leading to
surface-based descriptors, jump edges (edges separating the object
and background) and crease edges (edges between surfaces). These
representations naturally led to graph-based or structural matching
algorithms. Another approach based on interpretation trees yielded a
class of algorithms for object recognition. The theory of invariants
became popular with the goal of recognizing objects over large view
points. More information on these developments can be found in
\cite{Kanade_book, Grimson_book, Faugeras_book, Rosenfeld_book,
Horn_book, Szeliski_book}.

While these approaches were being developed, methods based on
artificial neural networks (ANNs) made a comeback in the
mid-eighties. The emergence of ANNs was largely motivated by the
excitement generated by Hopfield network's ability to address the
traveling salesman problem and the rediscovery of back-propagation
algorithm for training the ANNs. The first international conference
on neural networks held in 1987 in San Diego attracted lot more
attendees than expected. Firm believers in ANNs thought that a new
era had begun, and some even claimed that the demise of artificial
intelligence was inevitable!  The ANNs were not broadly welcomed by
computer vision researchers. The reason for the ambivalence of
computer vision researchers to ANNs was understandable. Computer
vision researchers were brought up with the notion that
representations derived from geometric, photometric as well as human
vision points of view were critical for the success of object
recognition systems ~\cite{Marr_book}. The approach of simply
feeding images into a 3-layer ANN and getting the labels out using
training data was not appealing to most computer vision researchers.
For one thing, it was not clear how general invariances could be
integrated into ANNs, despite early attempts of Fukushima in
designing the Neocognitron. ANN experiments that demonstrated object
recovery from partial information were not convincing.  ANN
researchers relied on the theory that the 3-layer networks can
represent the mapping between inputs and the class labels, given the
proper amount of neurons and training data. This claim did not
satisfy computer vision researchers as it was not clear how the
non-linear mappings in ANNs explicitly span the variations due to
pose, illumination, occlusion etc. Also, computer vision researchers
were more interested in 3D object recognition problems and were not
into OCRs where the ANNs were being applied. More information on
these developments can be found in \cite{Fukushima_1980,
Werbos_book, Rumelhart_book, Carpenter_book, Chellappa_book}.

While the ANNs were becoming popular, a class of networks known as
Convolutional Neural Networks (CNNs) was being developed by LeCun
and associates. The CNNs showed much promise in the domain of OCR.
The CNNs represented the idea that one can learn the representations
from data using learning algorithms. The tension between learning
representations directly from data vs handcrafting the
representations and applying appropriate preprocessing steps has
been ever present. The emergence of representations such as the
Scale-invariant feature transform (SIFT), which showed an order of
magnitude improvement when compared to interest points developed
more than three decades ago, the histogram of gradients (HoG)
operator and the local binary pattern \cite{ahonen_face_2006} are
good examples of hand-crafted features. CNNs left the feature
extraction work to a learning algorithm. Irrespective of whether
hand crafted or data-driven features extracted, there was a common
agreement on the effectiveness of support vector machines as
classifiers. More information on these developments can be found in
\cite{Farabet_PAMI2013 ,Vapnik_book, LeCun_cvpr2004, LeCun_1998,
LeCun_phd}.

Since the mid-nineties, computer vision researchers got interested
in problems such as video-based tracking, surveillance and activity
recognition, enabled by the availability of large volumes of video
data from stationary and moving cameras. Other problems such as
video stabilization, 3D modeling from multiple cameras, gait
analysis etc. also were pursued.  Another interesting development is
the willingness of researchers to be challenged with large data sets
and performance expectations. While these challenge problems were
first introduced in the OCR community, they soon made their way into
object detection and recognition (PASCAL VOC, ImageNet), face
recognition (FRGC, LFW, IJB-A, MegaFace) and activity recognition
research communities. It has become commonplace to have one or more
new challenge problems to be introduced every year. Many research
programs funded by the Government also introduced data sets and
evaluation protocols appropriate for measuring progress in their
programs. More detains on these developments can be found
\cite{Isard_IJCV, Zhao_survey, Nixon_book, Turaga_2008, CSVT_book,
Arnold_pami2014, Ikeuchi_book, FRGC, Everingham, Imagenet}.

The undaunted stalwarts of ANNs continued their quest for improving
the performance by increasing the number of layers. Since the
effectiveness of backpropagation algorithm was diminishing as the
number of layers increased, unsupervised methods based on Boltzman
machines \cite{Hinton} and autoencoders \cite{Bengio} were suggested
for obtaining good initial values of network parameters which were
then fed into deep ANNs.

\begin{figure*}[htp!]
\begin{center}
 \includegraphics[width=6.5in]{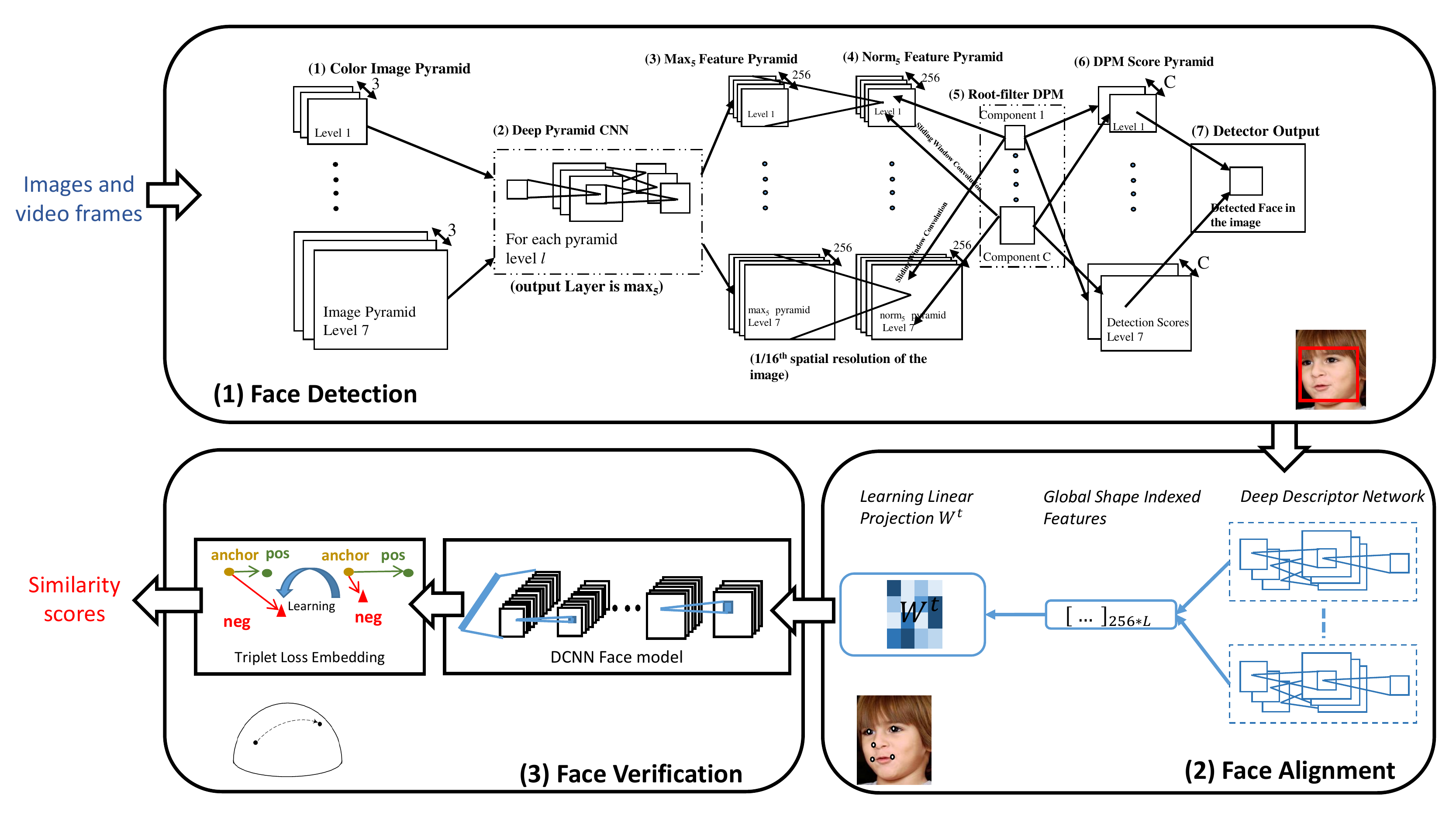}
\end{center}
  \caption{An overview of the proposed end-to-end DCNN-based face verification system \cite{Janus_ICCV_2015}.}
  \label{fig:system_overview}
\end{figure*}

As these developments were being made, the ``Eureka" moment came
about when DCNNs were first deployed for the ImageNet challenge a
mere four years back. The performance improvements obtained by DCNNs
\cite{NIPS2012_imagenet} for the ImageNet challenge were quite good.
The power of depth, the availability of GPUs and large annotated
data, replacement of traditional sigmoidal nonlinearities by
Rectified Linear Units  (ReLU) and drop out strategies were embodied
in the network now known as AlexNet \cite{NIPS2012_imagenet}. The
life of computer vision researchers has not been the same since!

The success of AlexNet motivated researchers from companies and
numerous universities  \cite{Deepface, Facenet, DeepID} to design
various versions of DCNNs by changing the number of layers, the
amount of training data being used and modifications to the
nonlinearities, etc. The tables shown below document the
improvements on object detection, image classification and face
verification over the last several years.

\begin{table}[htp!]
\centering
\begin{tabular}{|c|c|c|c|}
    \hline Rank & Name & Error rate  & Description \\
    \hline\hline
    1&Google&0.06656&Deep learning\\
    \hline
    2&Oxford&0.07325&Deep learning\\
    \hline
    3&MSRA&0.08062&Deep learning\\
    \hline
\end{tabular}
\caption{Top ImageNet 2014 image classification challenge results (\text{http://image-net.org/challenges/LSVRC/2014/}). }
\end{table}

\begin{table}[htp!]
\centering
\begin{tabular}{|c|c|c|c|}
    \hline Rank & Name & Error rate  & Description \\
    \hline\hline
1&Google&0.43933&Deep learning\\
\hline
2&CHUK&0.40656&Deep learning\\
\hline
3&DeepInsight&0.40452&Deep learning\\
\hline
4&UvA-Euvision&0.35421&Deep learning\\
\hline
5&Berkeley Vision&0.34521&Deep learning\\
    \hline
\end{tabular}
\caption{Top ImageNet 2014 object detection challenge results (\text{http://image-net.org/challenges/LSVRC/2014/}). }
\end{table}

\begin{table}[htp!]
\centering
\begin{tabular}{|c|c|c|c|}
    \hline Method & Result \\
    \hline\hline
DeepID2&    0.9915 $\pm$ 0.0013\\
TCIT&   0.9333 $\pm$ 0.0124\\
DeepID2+&   0.9947 $\pm$ 0.0012\\
betaface.com &  0.9808 $\pm$ 0.0016\\
DeepID3&    0.9953 $\pm$ 0.0010\\
insky.so&   0.9551 $\pm$ 0.0013\\
Uni-Ubi&    0.9900 $\pm$ 0.0032\\
FaceNet&    0.9963 $\pm$ 0.0009\\
Tencent-BestImage&  0.9965 $\pm$ 0.0025\\
Baidu&  0.9977 $\pm$ 0.0006\\
AuthenMetric&   0.9977 $\pm$ 0.0009\\
MMDFR&  0.9902 $\pm$ 0.0019\\
CW-DNA-1&   0.9950 $\pm$ 0.0022\\
Sighthound&     0.9979 $\pm$ 0.0003\\
    \hline
\end{tabular}
\caption{Top mean classification accuracy and standard error of the mean on the LFW dataset (\text{http://vis-www.cs.umass.edu/lfw/results.html}).}
\end{table}

While some may be dismayed by the reemergence of the so called
``black box" approach to computer vision problems such as object
detection/recognition and face verification/recognition, the simple
fact of life is that it is hard to argue against performance.  What
is comforting though, the original issues in object recognition
problems, such as robustness to pose, illumination variations,
degradations due to low-resolution, blur and occlusion still remain.
While using hundreds of million face images of millions subjects has
produced some of the best results in LFW face challenge, the
recently reported recognition performance of a DCNN with 500 million
faces of 10 million subjects in the low seventies for the MegaFace
challenge,  clearly argues for much more work to be done.

Over three decades, the author's group explored both traditional
computer vision approaches and ANN-based approaches for a variety of
problems in computer vision including object/face detection, face
verification/recognition.  An earlier attempt \cite{ATR_cnn} on
designing FLIR ATRs using CNNs undertaken by the author's group
yielded discouraging results as the number of layers used was
insufficient. Although sufficient training data was available for a
ten-class ATR problem, not having GPUs hampered the development of
deeper CNNs. Over the last eighteen months, under the support of
IARPA JANUS program for unconstrained face verification/recognition,
the authors have designed an end-to-end automatic face verification
and recognition system using DCNNs for face detection
\cite{ranjan_deep_2015}, face alignment \cite{Janus_ICCV_2015} and
verification/recognition stages \cite{Janus_ICCV_2015,
JC_WACV_2016}. This has been made possible due to the availability
of affordable GPU cards, large annotated face data sets and
implementations of CNNs such as Caffe \cite{jia_caffe_2014}, Torch,
CUDa-ConvNet, Theano and TensorFlow. Even high-school students are
able to build vision applications using DCNNs \cite{Age_ICCV_2015}!
While this is an exciting development for most researchers
interested in designing end-to-end object recognition system, we
feel we are at a stage similar to when Model-T was introduced many
decades ago.


\section{UMD Face Verification System} \label{sec:janus}
In this section, we give a brief overview of the UMD face verification system
\cite{JC_WACV_2016}, \cite{Janus_ICCV_2015} based on DCNNs.
Figure~\ref{fig:system_overview} shows the block diagram of the overall
verification system. First, faces are detected in each image and video frame.
Then, the detected faces are aligned into canonical coordinates using the
detected landmarks.  Finally, face verification is performed using the
low-dimensional features obtained using a triplet loss formulation to compute
the similarity between a pair of images or videos.  In what follows, we describe
each of these blocks in
detail.

\subsection{Face Detection}
\begin{figure}[htp!]
\begin{center}
 \includegraphics[width=2.6in]{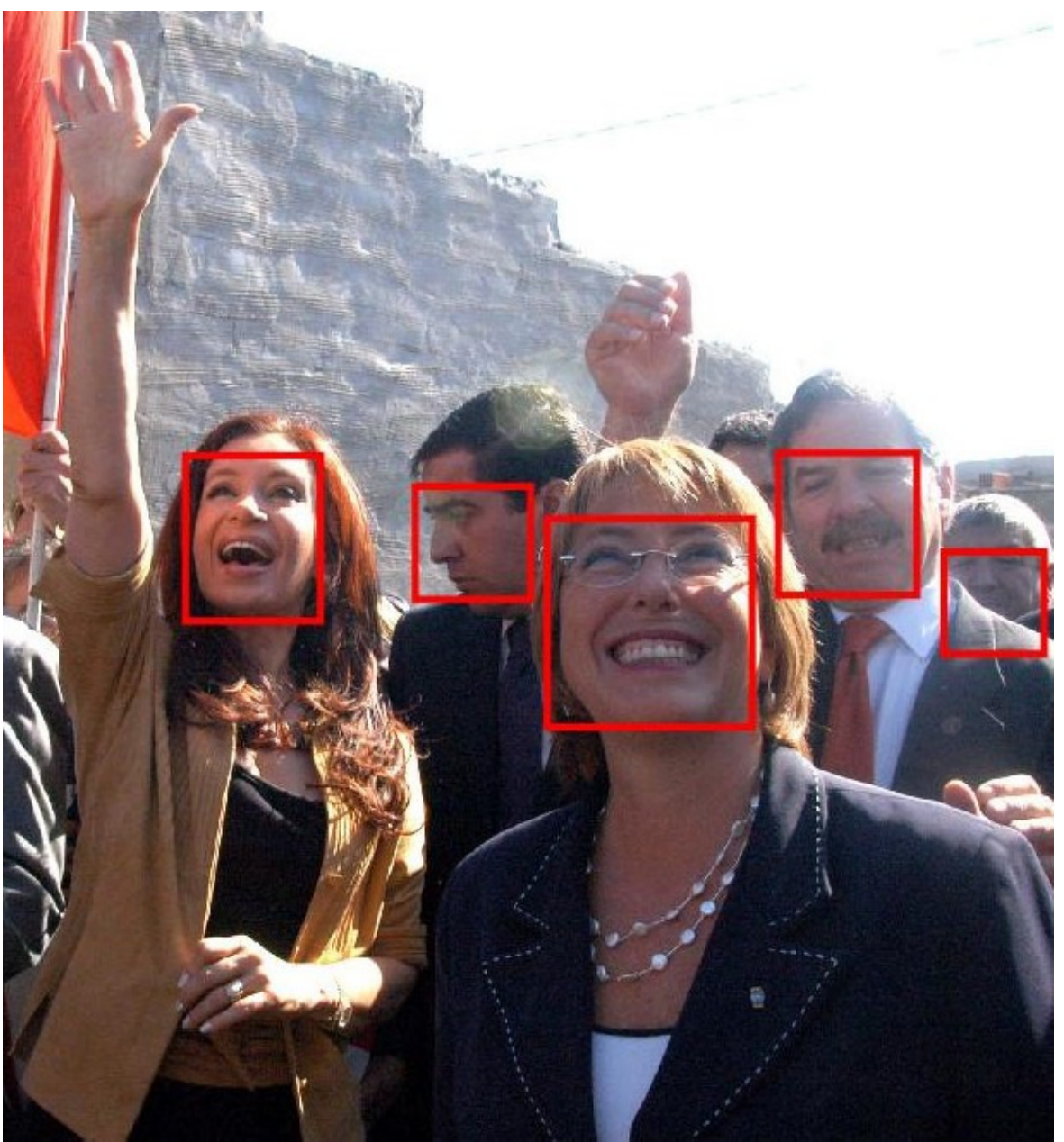}
\end{center}
  \caption{Sample detection results on an IJB-A image using the deep pyramid method.}
  \label{fig:detectedfaces}
\end{figure}

Faces in the images/videos are detected using the Deep Pyramid
Deformable Parts Model for Face Detection (DP2MFD) approach
presented in \cite{ranjan_deep_2015}. This method consists of two
main stages. The first stage generates a seven level normalized deep
feature pyramid for any input image of arbitrary size. The second
stage is a linear SVM, which takes these features as input to
classify each location as face or non-face, based on their scores.
It was shown in \cite{ranjan_deep_2015} that this detector is robust
to not only pose and illumination variations but also to different
scales. Furthermore, DP2MFD was shown to perform better than an
earlier version of Region with CNN (RCNN) based detector
\cite{girshick14CVPR}. Figure~\ref{fig:detectedfaces} shows some
detected faces by DP2MFD.


%

\subsection{Landmark Localization}
\begin{figure}[htp!]
\begin{center}
 \includegraphics[width=3in]{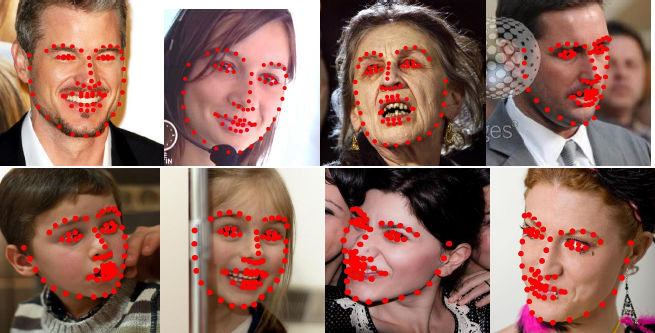}
\end{center}
  \caption{Sample facial landmark detection results \cite{Janus_ICCV_2015}.}
  \label{fig:face_landmark}
\end{figure}

In the second step of our verification system, facial landmarks are
detected to align the detected faces.  The landmark detection
problem can be viewed as a regression problem, where  beginning with
the initial mean shape, the target shape is reached through
regression. First, deep features are extracted of a patch around a
point of the shape.  Then, the cascade regression in which the
output generated by the first stage is used as an input for the next
stage is used to learn the regression function. After the facial
landmark detection is completed, each face is aligned into the
canonical coordinate with similarity transform using the 3 landmark
points (i.e. the center of each eye and the base of the nose). After
alignment, the face image resolution is 100 $\times$ 100 pixels.
Examples of detected landmarks results are shown in
Figure~\ref{fig:face_landmark}.

\subsection{Face Verification}
Once the faces are detected and aligned, we train a DCNN network using the
architecture presented in \cite{JC_WACV_2016}.  The network is trained using the
CASIA-WebFace dataset \cite{yi_learning_2014}. The dimensionality of the input
layer is $100 \times 100 \times 3$ for gray-scale images. The network includes
10 convolutional layers, 5 pooling layers, and 1 fully connected layer.
The final dimensionality of the feature is 320.   More details of the DCNN
network can be found in \cite{JC_WACV_2016}.

To further improve the performance of our deep features, we obtain a
low-dimensional discriminative projection of the deep features that
is learnt using the training data provided for each split of IJB-A
(section D below describes the IJB-A data set). The output of the
procedure is an embedding matrix $\mathbf{W} \in \mathbf{R}^{M
\times n}$ where $M$ is the dimensionality of the deep descriptor,
which is 320 in our case and we set $n=128$, thus achieving
dimensionality reduction in addition to an improvement in
performance.

The objective of this method is to push similar pairs together and
dissimilar pairs apart in the low-dimensional space. For learning
$\mathbf{W}$, we solve an optimization problem based on constraints
involving triplets - each containing two similar samples and one
dissimilar sample. The triplets are generated using the label
information provided with the training data for each split. This
optimization is solved using stochastic gradient descent and the
entire procedure takes 3-5 minutes per split using a standard C++
implementation.

\subsection{Experimental Results}
To show the performance of the UMD DCNN-based face verification system, we
highlight some verification results on the challenging IARPA JANUS Benchmark A
(IJB-A) ~\cite{klare_janus_2015} and its extended version JANUS Challenging set
2 (JANUS CS2) dataset \cite{Janus_ICCV_2015}.  The JANUS CS2 dataset contains
not only the sampled frames and images in the IJB-A, but also the original
videos. In addition, the JANUS CS2 dataset\footnote{The JANUS CS2 dataset is not
publicly available yet.} includes considerably more test data for identification
and verification problems in the defined protocols than the IJB-A dataset.
In both the IJB-A and the CS2 data set the verification and recognition
protocols are set around templates, which are groups of one or more images of
the same person. Verification is template to template, recognition is template
to gallery (of templates).

Table~\ref{exp:roc_cmc_scores_ijba} summarizes the scores produced
by different face verification methods on the IJB-A dataset.  Among
the methods compared in Table~\ref{exp:roc_cmc_scores_ijba} include,
a DCNN-based method \cite{wang_face_2015}, Fisher vector-based
method \cite{simonyan_fisher_2013}, and one commercial off-the-shelf
matcher, COTS \cite{klare_janus_2015} which are tested in a fully
automatic setup. The DCNN that performs verification uses data
processed by the automated face preprocessing components described
in previous subsections. This is followed by fine-tuning, triplet
embedding, and testing steps.

Since the system works end-to-end, we have devised two ways of handling
the situation where we are unable to detect any of the faces in the images
of given template:
\begin{itemize}
  \item \textbf{Setup 1:} Under this setup, our verification and recognition accuracy is
    measured only over images we can process.  The philosophy for this setup
    is: assume average performance for any images we cannot process.  For
    verification: if in a template to template comparison in which we cannot
    detect a face in any of its images, we ignore this template to template
    comparison. For recognition: if we cannot process any of the images of an
    individual to compare versus a gallery we do not include this probe in the
    computation of recognition rates.
  \item \textbf{Setup 2:} Under this setup, our verification and recognition
    accuracy is measured in a pessimistic fashion: we are forced to make a
    decision even if we have not been able to detect faces. In this case, for
    verification we set the score to the lowest possible similarity (when we're
    unable to process any of the images in one of the two templates being
    compared).  For recognition experiments, we set the rank as the highest
    possible rank, when we're unable to process all the images in the probe
    template.
  \item \textbf{Setup 3:} In this setup, we include our earlier results ~\cite{Janus_ICCV_2015}
    for the purpose of comparison.
\end{itemize}

   The following lists the differences of the components used in the current
   work from \textbf{Setup 3}:
   \begin{itemize}
   \item We adopt the same network architecture presented in ~\cite{Janus_ICCV_2015}; However, for the training data,
     we use RGB color and larger face region (\emph{i.e.}, crop 125 $\times$
     125-pixel face regions and resize them to 100 $\times$ 100 pixels.) during
     alignment instead of gray scale and smaller face regions.
   \item We replace the joint Bayesian metric with triplet loss embedding.
   \item Complete implementation in C++ instead of MATLAB.
   \item \textbf{Setup 3} has one extra component, face association, to
     detect and track faces across frames. The association component requires
     one of ground truth face bounding boxes as the initialization to specify
     which face track to be used for comparison. Thus, if all the faces are not
     detected within a template, it still has one face for comparison as
     compared to other setups. (\emph{i.e.} However, the faces are difficult
     ones, and most of them are in extreme pose and illumination variations.
     Thus, including these faces does not improve the overall performance a
     lot.)
\end{itemize}

\begin{table*}[htp!]
\centering \small
\begin{tabular}{|c|c|c|c|c|}
  \hline
  IJB-A-Verif& \cite{wang_face_2015} & DCNN (setup 1) & DCNN (setup 2) & DCNN (setup 3)\\
  \hline
  FAR=1e-2&0.732 $\pm$ 0.033&0.8312 $\pm$ 0.0350&0.7810 $\pm$ 0.0316 & 0.776 $\pm$ 0.033  \\
  FAR=1e-1&0.895 $\pm$ 0.013&0.9634 $\pm$ 0.0049&0.9006 $\pm$ 0.0077& 0.936 $\pm$ 0.01 \\
  \hline\hline
  IJB-A-Ident & \cite{wang_face_2015} & DCNN (setup 1) & DCNN (setup 2) & DCNN (setup 3) \\
  \hline
  Rank-1 &0.820 $\pm$ 0.024&0.8990 $\pm$ 0.0105&0.8378 $\pm$ 0.0142 & 0.834 $\pm$ 0.017\\
  Rank-5 &0.929 $\pm$ 0.013&0.9706 $\pm$ 0.0075&0.9073 $\pm$ 0.0119 & 0.922 $\pm$ 0.011 \\
  Rank-10&$N/A$&0.9821 $\pm$ 0.0053&0.9219 $\pm$ 0.0094 & 0.947 $\pm$ 0.011  \\
  \hline
\end{tabular}
 \vspace{+1mm} \caption{Results on the IJB-A
dataset. The TAR of all the approaches at FAR=0.1 and 0.01 for the
ROC curves. The Rank-1, Rank-5, and Rank-10 retrieval accuracies of
the CMC curves. We report average and standard deviation of the 10 splits.}
\label{exp:roc_cmc_scores_ijba}
\end{table*}

\begin{figure*}[htp!]
\centering \subfigure[]{
\includegraphics[height=2.5in]{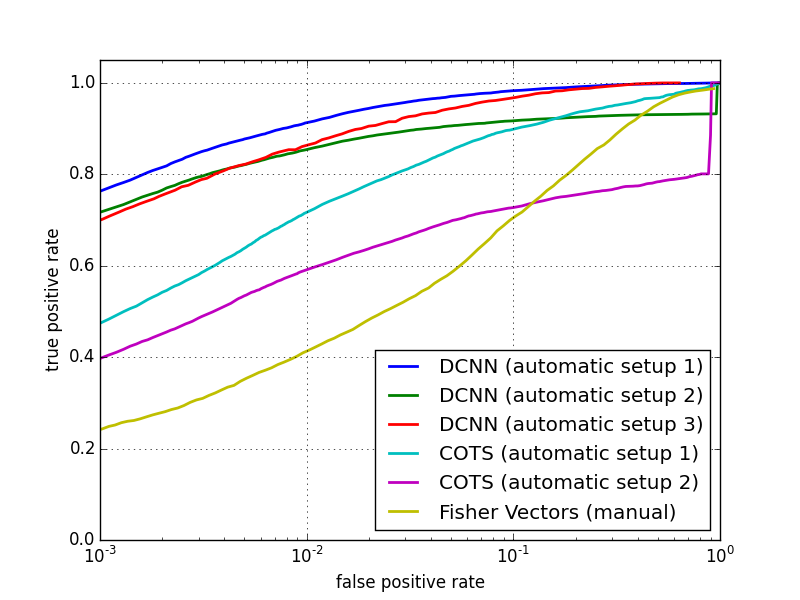}
} \subfigure[]{
\includegraphics[height=2.5in]{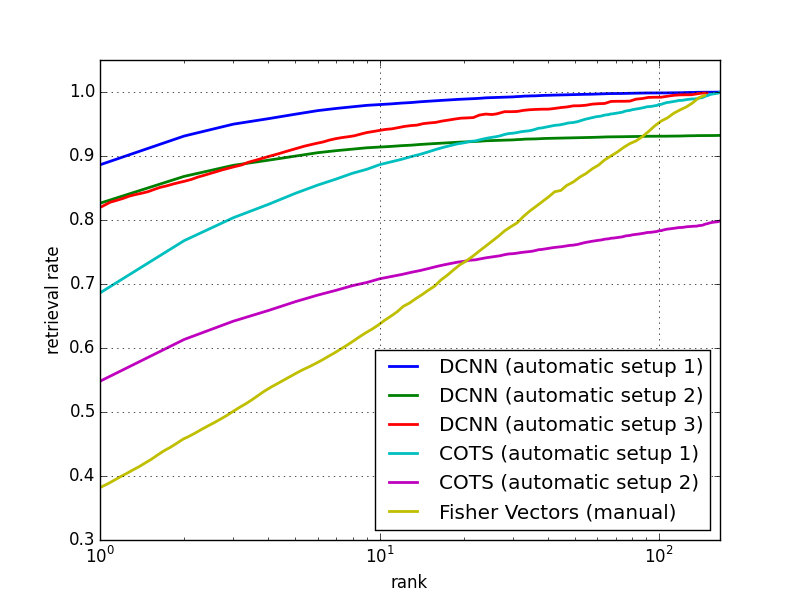}
} \vspace{-1mm}  \caption{Results on the JANUS CS2 dataset.  (a) the
average ROC curves and (b) the average CMC
curves. Our curves are the average curve of the 10 splits.}\label{exp:janus_roc_cmc}
\end{figure*}

Figure~\ref{exp:janus_roc_cmc} shows the ROC curves and the CMC curves
corresponding to different methods on the JANUS CS2 dataset, respectively for
verification and identification protocols.  The corresponding scores are
summarized in Table~\ref{exp:roc_cmc_scores1}.   From the ROC and CMC curves, we
see that the DCNN method performs better than other competitive methods. This
can be attributed to the fact that the DCNN model does capture face variations
over a large dataset and generalizes well to a new small dataset. In addition,
the performance of the proposed automatic system degrades only slightly as
compared to the one using the manual annotations. This demonstrates the
robustness of each component of our system.

\begin{table*}[htp!]
\centering \small
\begin{tabular}{|c|c|c|c|c|c|c|}
  \hline
  JANUS-CS2-Verif&COTS (setup 1)&COTS (setup 2)&FV\cite{simonyan_fisher_2013}&DCNN (setup 1)&DCNN (setup 2) & DCNN (setup 3)\\
  \hline
  FAR=1e-2&0.7167 $\pm$ 0.0168&0.5917 $\pm$ 0.0154&0.411 $\pm$ 0.081&0.9132 $\pm$ 0.0059&0.8538 $\pm$ 0.008 & 0.861 $\pm$ 0.014\\
  FAR=1e-1&0.8986 $\pm$ 0.0099&0.7272 $\pm$ 0.0133&0.704 $\pm$ 0.028&0.9829 $\pm$ 0.0038&0.9170 $\pm$ 0.0074 & 0.963 $\pm$ 0.007\\
  \hline\hline
  JANUS-CS2-Ident&COTS (setup 1)&COTS (setup 2)&FV\cite{simonyan_fisher_2013}&DCNN (setup 1)&DCNN (setup 2)& DCNN (setup 3)\\
  \hline
  Rank-1&0.6861 $\pm$ 0.0156&0.5481 $\pm$ 0.0150&0.381 $\pm$ 0.018&0.8862 $\pm$ 0.0089&0.8262 $\pm$ 0.0100 & 0.82 $\pm$ 0.014\\
  Rank-5&0.8417 $\pm$ 0.0111&0.6724 $\pm$ 0.0141&0.559 $\pm$ 0.021&0.9653 $\pm$ 0.0057&0.9000 $\pm$ 0.0086 & 0.91 $\pm$ 0.01\\
  Rank-10&0.8865 $\pm$ 0.0091&0.7081 $\pm$ 0.0125&0.637 $\pm$ 0.025&0.9803 $\pm$ 0.0047&0.9139 $\pm$ 0.0081 & 0.938 $\pm$ 0.01\\
  \hline
\end{tabular}
\vspace{+1mm} \caption{Results on the JANUS CS2 dataset. The TAR of
all the approaches at FAR=0.1 and 0.01 for the ROC curves. The
Rank-1, Rank-5, and Rank-10 retrieval accuracies of the CMC curves. We report
average and standard deviation of the 10 splits.}
\label{exp:roc_cmc_scores1}
\end{table*}

\section{Open Issues}\label{sec:con}
Given sufficient number of annotated data and GPUs, DCNNs have been shown to
yield impressive performance improvements.  Still many issues remain to be
addressed to make the DCNN-based recognition systems robust and practical. These
are briefly discussed below.

\begin{itemize}
\item {\textbf{Reliance on large training data sets:}}  As discussed before, one of the
  top performing networks in the MegaFace challenge needs 500 million faces of about 10 million subjects. Such
  large annotated training set may not be always available (e.g. expression
  recognition, age estimation). So networks that can perform well with
  reasonable-sized training data are needed.

\item {\textbf{Invariance:}} While limited invariance to translation is possible
  with existing DCNNs, networks that can incorporate more general 3D invariances
  are needed.

\item {\textbf{Training time:}} The training time even when GPUs are used can be
  several tens to hundreds of hours, depending on the number of layers used and
  the training data size. More efficient implementations of learning algorithms,
  preferably implemented using CPUs are desired.

\item {\textbf{Number of parameters:}}  The number of parameters can be several
  tens of millions. Novel strategies that reduce the number of parameters need
  to be developed.

\item {\textbf{Handling degradations in training data:}} : DCNNs robust to
  low-resolution, blur, illumination and pose variations, occlusion, erroneous annotation,  etc. are needed to
  handle degradations in data.

\item {\textbf{Domain adaptation of DCNNs:}} While having large volumes of data
  may help with processing test data from a different distribution than that of
  the training data, systematic methods for adapting the deep features to test
  data are needed.

\item {\textbf{Theoretical considerations:}} While DCNNs have been around for a
  few years, detailed theoretical understanding is just starting to develop ~\cite{bruna_invariant_2013, mallat_understanding_2016, Raja_deep, vidal_deep}.
  Methods for deciding the number of layers, neighborhoods over which max pooling operations
  are performed are needed.

\item {\textbf{Incorporating domain knowledge:}} The current practice is to rely
  on fine tuning. For example, for the age estimation problem, one can start
  with one of the standard networks such as the AlexNet and fine tune it using
  aging data. While this may be reasonable for somewhat related problems (face
  recognition and facial expression recognition), such fine tuning strategies
  may not always be effective. Methods that can incorporate context may make the
  DCNNs more applicable to a wider variety of problems.

\item {\textbf{Memory:}} The DCNNs in use currently are memoryless, in that they
  cannot predict based on the data they have seen before. This reduces their
  ability to process videos. Although Recurrent CNNs are on the rise, they
  still consume a lot of time and memory for training and deployment.
  Efficient DCNN algorithms are needed to handle videos and
  other data streams as blocks.

\end{itemize}

\section*{Acknowledgments}
This research is based upon work supported by the Office of the Director of
National Intelligence (ODNI), Intelligence Advanced Research Projects Activity
(IARPA), via IARPA R\&D Contract No. 2014-14071600012. The views and conclusions
contained herein are those of the authors and should not be interpreted as
necessarily representing the official policies or endorsements, either expressed
or implied, of the ODNI, IARPA, or the U.S. Government. The U.S. Government is
authorized to reproduce and distribute reprints for Governmental purposes
notwithstanding any copyright annotation thereon.

\bibliographystyle{ieee}
\bibliography{ITA,refs_ching}

\begin{thebibliography}{10}\itemsep=-1pt

\bibitem{ahonen_face_2006}
T.~Ahonen, A.~Hadid, and M.~Pietikainen.
\newblock Face description with local binary patterns: Application to face
  recognition.
\newblock {\em {IEEE} Transactions on Pattern Analysis and Machine
  Intelligence}, 28(12):2037--2041, 2006.

\bibitem{Bengio}
Y.~Bengio.
\newblock Learning deep architectures for ai.
\newblock {\em Found. Trends Mach. Learn.}, 2(1):1--127, Jan. 2009.

\bibitem{bruna_invariant_2013}
J.~Bruna and S.~Mallat.
\newblock Invariant scattering convolution networks.
\newblock {\em IEEE Transactions on Pattern Analysis and Machine Intelligence},
  35(8):1872--1886, 2013.

\bibitem{Carpenter_book}
G.~Carpenter and S.~Grossberg.
\newblock {\em Pattern Recognition by Self-Organizing Neural Networks}.
\newblock A Bradford Book, 1991.

\bibitem{CSVT_book}
R.~Chellappa, A.~Sankaranarayanan, A.~Veerraghavan, and P.~Turaga.
\newblock {\em Statistical Models and Methods for Video-based Tracking and
  Recognition}.
\newblock Now Publishers Inc, 2010.

\bibitem{JC_WACV_2016}
J.-C. Chen, V.~M. Patel, and R.~Chellappa.
\newblock Unconstrained face verification using deep cnn features.
\newblock In {\em IEEE Winter conference on Applications of Computer Vision},
  2016.

\bibitem{Janus_ICCV_2015}
J.-C. Chen, R.~Ranjan, A.~Kumar, C.-H. Chen, V.~M. Patel, and R.~Chellappa.
\newblock An end- to-end system for unconstrained face verification with deep
  convolutional neural networks.
\newblock In {\em IEEE International Conference on Computer Vision (ICCV)
  workshop on ChaLearn Looking at People (ChaLearn LaP)}, 2016.

\bibitem{Imagenet}
J.~Deng, W.~Dong, R.~Socher, L.-J. Li, K.~Li, and L.~Fei-Fei.
\newblock Imagenet: A large-scale hierarchical image database.
\newblock In {\em IEEE Conference on Computer Vision and Pattern Recognition},
  pages 248--255, June 2009.

\bibitem{Everingham}
M.~Everingham, L.~V. Gool, C.~K. I.Williams, J.Winn, and A.~Zisserman.
\newblock The {PASCAL} visual object classes {(VOC)} challenge.
\newblock {\em International Journal of Computer Vision}, 88(2):303--338, June
  2010.

\bibitem{Farabet_PAMI2013}
C.~Farabet, C.~Couprie, L.~Najman, and Y.~LeCun.
\newblock Learning hierarchical features for scene labeling.
\newblock {\em IEEE Transactions on Pattern Analysis and Machine Intelligence},
  35(8):1915--1929, Aug 2013.

\bibitem{Faugeras_book}
O.~Faugeras.
\newblock {\em Three-dimensional Computer Vision: A Geometric View Point}.
\newblock The MIT Press, 1993.

\bibitem{fu_book}
K.~S. Fu.
\newblock {\em Syntactic Pattern Recognition and Applications}.
\newblock Prentice-Hall, 1981.

\bibitem{Fukunaga_book}
K.~Fukunaga.
\newblock {\em Introduction to Statistical Pattern Recognition}.
\newblock Academic Press, 1990.

\bibitem{Fukushima_1980}
K.~Fukushima.
\newblock Neocognitron: A self-organizing neural network model for a mechanism
  of pattern recognition unaffected by shift in position.
\newblock {\em Biological Cybernetics}, 36(4):93--202, 1980.

\bibitem{girshick14CVPR}
R.~Girshick, J.~Donahue, T.~Darrell, and J.~Malik.
\newblock Rich feature hierarchies for accurate object detection and semantic
  segmentation.
\newblock In {\em Computer Vision and Pattern Recognition (CVPR), 2014 IEEE
  Conference on}, pages 580--587, 2014.

\bibitem{Raja_deep}
R.~Giryes, G.~Sapiro, and A.~M. Bronstein.
\newblock On the stability of deep networks.
\newblock {\em {arXiv} preprint {arXiv:1412.5896}}, 2014.

\bibitem{Grimson_book}
W.~Grimson.
\newblock {\em Object Recognition by Computer: The Role of Geometric
  Constraints}.
\newblock MIT Press, Cambridge, MA, 1990.

\bibitem{vidal_deep}
B.~D. Haeffele and R.~Vidal.
\newblock Global optimality in tensor factorization, deep learning, and beyond.
\newblock {\em {arXiv} preprint {arXiv:1506.07540}}, 2015.

\bibitem{Hinton}
G.~E. Hinton and T.~J. Sejnowski.
\newblock {\em Learning and relearning in Boltzmann machines in Parallel
  Distributed Processing: Explorations in the Microstructure of Cognition.
  Volume 1: Foundations}.
\newblock MIT Press, Cambridge, MA, 1986.

\bibitem{Horn_book}
B.~K.~P. Horn.
\newblock {\em Robot Vision}.
\newblock The MIT Press, Cambridge, MA, 1986.

\bibitem{Ikeuchi_book}
K.~Ikeuchi.
\newblock {\em Computer Vision: A Reference Guide}.
\newblock Springer, 2014.

\bibitem{Isard_IJCV}
M.~Isard and A.~Blake.
\newblock Condensation-conditional density propagation for visual tracking.
\newblock {\em International Journal on Computer Vision}, 29(1):5--28, 1998.

\bibitem{jia_caffe_2014}
Y.~Jia, E.~Shelhamer, J.~Donahue, S.~Karayev, J.~Long, R.~Girshick,
  S.~Guadarrama, and T.~Darrell.
\newblock Caffe: Convolutional architecture for fast feature embedding.
\newblock In {\em ACM International Conference on Multimedia}, pages 675--678,
  2014.

\bibitem{Kanade_book}
T.~Kanade.
\newblock {\em Three Dimensional Vision}.
\newblock Kluwer Academic Publishers, Boston, MA, 1987.

\bibitem{klare_janus_2015}
B.~F. Klare, B.~Klein, E.~Taborsky, A.~Blanton, J.~Cheney, K.~Allen,
  P.~Grother, A.~Mah, M.~Burge, and A.~K. Jain.
\newblock Pushing the frontiers of unconstrained face detection and
  recognition: {IARPA Janus Benchmark A}.
\newblock In {\em IEEE Conference on Computer Vision and Pattern Recognition},
  2015.

\bibitem{NIPS2012_imagenet}
A.~Krizhevsky, I.~Sutskever, and G.~E. Hinton.
\newblock Imagenet classification with deep convolutional neural networks.
\newblock In F.~Pereira, C.~Burges, L.~Bottou, and K.~Weinberger, editors, {\em
  Advances in Neural Information Processing Systems 25}, pages 1097--1105.
  Curran Associates, Inc., 2012.

\bibitem{LeCun_phd}
Y.~LeCun.
\newblock {\em Mod`eles connexionistes de lÕapprentissage}.
\newblock PhD thesis, Universit«e de Paris VI, 1987.

\bibitem{LeCun_1998}
Y.~LeCun, L.~Bottou, Y.~Bengio, and P.~Haffner.
\newblock Gradient-based learning applied to document recognition.
\newblock {\em Proceedings of the IEEE}, 86(11):2278--2324, 1998.

\bibitem{LeCun_cvpr2004}
Y.~LeCun, F.-J. Huang, and L.~Bottou.
\newblock Learning methods for generic object recognition with invariance to
  pose and lighting.
\newblock In {\em IEEE Conference on Computer Vision and Pattern Recognition},
  pages 97--104, 2004.

\bibitem{ATR_cnn}
B.~Li, R.~Chellappa, Q.~Zheng, S.~Der, N.~M. Nasrabadi, L.~Chan, and L.~Wang.
\newblock Experimental evaluation of flir atr approaches - a comparative study.
\newblock {\em Computer Vision and image understanding}, 84:5--24, 2001.

\bibitem{mallat_understanding_2016}
S.~Mallat.
\newblock Understanding deep convolutional networks.
\newblock {\em arXiv preprint arXiv:1601.04920}, 2016.

\bibitem{Marr_book}
D.~Marr.
\newblock {\em Vision}.
\newblock The MIT Press, Cambridge, MA, 1982.

\bibitem{Nixon_book}
M.~Nixon, T.~Tan, and R.~Chellappa.
\newblock {\em Human Identification Based on Gait}.
\newblock Springer, 2005.

\bibitem{FRGC}
P.~Phillips, P.~Flynn, T.~Scruggs, K.~Bowyer, J.~Chang, K.~Hoffman, J.~Marques,
  J.~Min, and W.~Worek.
\newblock Overview of the face recognition grand challenge.
\newblock In {\em IEEE Conference on Computer Vision and Pattern Recognition},
  volume~1, pages 947--954 vol. 1, June 2005.

\bibitem{ranjan_deep_2015}
R.~Ranjan, V.~M. Patel, and R.~Chellappa.
\newblock A deep pyramid deformable part model for face detection.
\newblock In {\em {IEEE} International Conference on Biometrics: Theory,
  Applications and Systems}, 2015.

\bibitem{Age_ICCV_2015}
R.~Ranjan, S.~Zhou, J.-C. Chen, A.~Kumar, A.~Alavi, V.~M. Patel, and
  R.~Chellappa.
\newblock Unconstrained age estimation with deep convolutional neural networks.
\newblock In {\em IEEE International Conference on Computer Vision (ICCV)
  workshop on ChaLearn Looking at People (ChaLearn LaP)}, 2016.

\bibitem{Rosenfeld_book}
A.~Rosenfeld and A.~Kak.
\newblock {\em Digital Picture Processing Vol 1 and 2}.
\newblock Academic Press, 1982.

\bibitem{Rumelhart_book}
D.~E. Rumelhart, J.~L. McClelland, and the PDP Research~Group.
\newblock {\em Parallel Distributed Processing: Explorations in the
  Microstructure of Cognition, Vol 1}.
\newblock Cambridge: MIT Press, 1986.

\bibitem{Facenet}
F.~Schroff, D.~Kalenichenko, and J.~Philbin.
\newblock Facenet: A unified embedding for face recognition and clustering.
\newblock In {\em IEEE Conference on Computer Vision and Pattern Recognition},
  pages 815--823, June 2015.

\bibitem{simonyan_fisher_2013}
K.~Simonyan, O.~M. Parkhi, A.~Vedaldi, and A.~Zisserman.
\newblock Fisher vector faces in the wild.
\newblock In {\em British Machine Vision Conference}, volume~1, page~7, 2013.

\bibitem{Arnold_pami2014}
A.~Smeulders, D.~Chu, R.~Cucchiara, S.~Calderara, A.~Dehghan, and M.~Shah.
\newblock Visual tracking: An experimental survey.
\newblock {\em IEEE Transactions on Pattern Analysis and Machine Intelligence},
  36(7):1442--1468, July 2014.

\bibitem{DeepID}
Y.~Sun, X.~Wang, and X.~Tang.
\newblock Deeply learned face representations are sparse, selective, and
  robust.
\newblock In {\em IEEE Conference on Computer Vision and Pattern Recognition},
  pages 2892--2900, June 2015.

\bibitem{Szeliski_book}
R.~Szeliski.
\newblock {\em Computer Vision: Algorithms and Applications}.
\newblock Springer, 2010.

\bibitem{Deepface}
Y.~Taigman, M.~Yang, M.~Ranzato, and L.~Wolf.
\newblock Deepface: Closing the gap to human-level performance in face
  verification.
\newblock In {\em IEEE Conference on Computer Vision and Pattern Recognition},
  pages 1701--1708, June 2014.

\bibitem{Turaga_2008}
P.~Turaga, R.~Chellappa, V.~Subrahmanian, and O.~Udrea.
\newblock Machine recognition of human activities: A survey.
\newblock {\em IEEE Transactions on Circuits and Systems for Video technology},
  18:1473--1488, Nov. 2008.

\bibitem{Vapnik_book}
V.~N. Vapnik.
\newblock {\em The Nature of Statistical Learning Theory}.
\newblock Springer, 1995.

\bibitem{wang_face_2015}
D.~Wang, C.~Otto, and A.~K. Jain.
\newblock Face search at scale: 80 million gallery.
\newblock {\em arXiv preprint arXiv:1507.07242}, 2015.

\bibitem{Werbos_book}
P.~Werbos.
\newblock {\em The roots of backpropagation: from ordered derivatives to neural
  networks and political forecasting}.
\newblock Wiley-Interscience, New York, NY, 1994.

\bibitem{yi_learning_2014}
D.~Yi, Z.~Lei, S.~Liao, and S.~Z. Li.
\newblock Learning face representation from scratch.
\newblock {\em arXiv preprint arXiv:1411.7923}, 2014.

\bibitem{Zhao_survey}
W.~Zhao, R.~Chellappa, J.~Phillips, and A.~Rosenfeld.
\newblock Face recognition in still and video images: A literature surrey.
\newblock {\em ACM Computing Surveys}, 35:399--458, Dec. 2003.

\bibitem{Chellappa_book}
Y.~Zhou and R.~Chellappa.
\newblock {\em Artificial Neural Networks for Computer Vision}.
\newblock Springer-Verlag, 1991.

\end{thebibliography}

\end{document}